# PARAMETER TUNING STRATEGIES FOR METAHEURISTIC METHODS APPLIED TO DISCRETE OPTIMIZATION OF STRUCTURAL DESIGN

Iván Negrin*[1], Dirk Roose** and Ernesto Chagoyén*

* Central University of Las Villas, Santa Clara, Cuba

** KU Leuven, Leuven, Belgium

[1] indiaz@uclv.cu

**ABSTRACT**

This paper presents several strategies to tune the parameters of metaheuristic methods for (discrete) design optimization of reinforced concrete (RC) structures. A novel utility metric is proposed, based on the area under the average performance curve. The process of modelling, analysis and design of realistic RC structures leads to objective functions for which the evaluation is computationally very expensive. To avoid costly simulations, two types of surrogate models are used. The first one consists of the creation of a database containing all possible solutions. The second one uses benchmark functions to create a discrete sub-space of them, simulating the main features of realistic problems. Parameter tuning of four metaheuristics is performed based on two strategies. The main difference between them is the parameter control established to perform partial assessments. The simplest strategy is suitable to tune good "generalist" methods, i.e., methods with good performance regardless the parameter configuration. The other one is more expensive, but is well suited to assess any method. Tuning results prove that Biogeography-Based Optimization, a relatively new evolutionary algorithm, outperforms other methods such as GA or PSO for such optimization problems, due to its particular approach of applying recombination and mutation operators.

**KEYWORDS:** Parameter Tuning, Metaheuristic Methods, Discrete Optimization

**RESUMEN**

Este artículo presenta varias estrategias para ajustar los parámetros de los métodos metaheurísticos en la optimización (discreta) del diseño estructural de estructuras de Hormigón Armado (HA). Una novedosa métrica de utilidad es propuesta, utilizando el área bajo la curva de rendimiento promedio. El proceso de modelación, análisis y diseño estructural de estructuras reales de HA deriva en funcione objetivos cuyas evaluaciones son muy costosas desde el punto de vista computacional. Para evitar estas costosas simulaciones, se emplean dos tipos de modelos sustitutivos. El primero consiste en la creación de una base de datos con todas las posibles soluciones. El segundo utiliza funciones de referencia para crear un sub-espacio discreto de estas, simulando las principales características de los problemas reales. Se ajustan los parámetros de cuatro metaheurísticas basado en dos estrategias. La mayor diferencia entre estas es el control de parámetros establecido para realizar mediciones parciales. La estrategia más simple es apropiada para ajustar métodos que sean buenos "generalistas", es decir, con buenos resultados sin importar la configuración de parámetros. La otra es más costosa, pero es útil para medir el rendimiento de cualquier método. Los resultados demostraron que la Optimización Basada en Biogeografía, un algoritmo evolutivo relativamente nuevo, supero en rendimiento a otros métodos como GA y PSO para estos problemas de optimización, debido a su particular enfoque de aplicar los operadores de recombinación y mutación.

**Palabras claves:** Ajuste de Parámetros, Métodos Metaheurísticos, Optimización Discreta

# 1. INTRODUCTION

Optimization is a topic that has been deeply studied in recent years due to the continuous development of computational tools and the need to optimize processes in order to use resources in a more efficient way.

To solve optimization problems, a large number of methods have been developed, depending on the type of problem to be solved and its characteristics (type and number of objective functions, types of variables). In the context of structural engineering, the design optimization of civil engineering structures has been studied by many researchers. In order to obtain results that are practical from an engineering point of view, this usually leads to a discrete and combinatorial optimization problem, for which the use of metaheuristics is the most common approach. These methods apply stochastic operators to explore the solution space and to guide the search towards optimal designs based on the objective(s) [1].

These operators and the parameters of the method can be tuned for the optimization problem at hand. The performance of a metaheuristic method depends on three main aspects: (1) the optimization problem, (2) the values of the parameters, and (3) the random variability inherent to stochastic algorithms [1]. Therefore, one of the challenges to solve an optimization problem is, in addition to find an appropriate method, to find a good configuration of its parameters. In case one is "testing" several methods to solve certain problems using bad parameter configurations, the results can lead to the wrong selection of the method. Therefore, a poor selection of its parameters could lead to a lack of robustness of the results [2] (see Fig. 1).

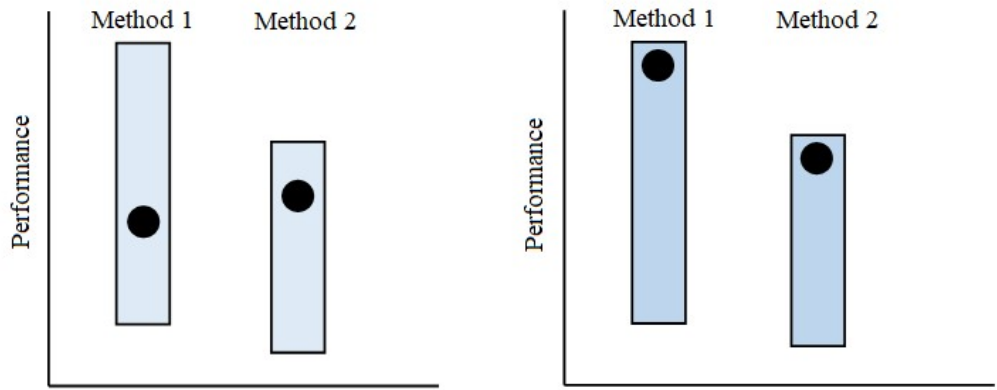

**Fig.1** The effect of parameter tuning on comparing metaheuristics. Left: the traditional situation, where the reported method performance is an "accidental" point on the scale ranging from the worst to the best performance (as determined by the parameter values used). Right: the improved situation, where the reported method performance is a near-optimal point on this scale, belonging to the tuned instance. This indicates the full potential of the given method, i.e., how good it can be when using the right parameter values [2].

According to [3], one can distinguish two approaches to improve methods' performance:

- Parameter tuning, where (good) parameter values are established before the run of a given metaheuristic. In this case, parameter values are fixed in the initialization stage and do not change while the method is running.
- Parameter control, where (good) parameter values are established during the run of a given metaheuristic. In this case, parameter values are given an initial value when starting the metaheuristic and they undergo changes while the method is running.

In our case, we are dealing with parameter tuning. However, the concept of parameter control can be applied to enhance the tuning procedure by means of pauses to check the method's performance, changing the initial configuration of parameter and parameter values depending on the analysis of preliminary results.

On the other hand, structural optimization processes are usually computationally expensive, therefore performing these tuning procedures is a difficult task. A common approach is to perform several independent runs of the algorithm for a given problem and to summarize the results using descriptive statistics [1]. The performance metrics (or utilities) commonly used in metaheuristics are: MBF (Mean Best Fitness), AES (average number of evaluations to solution) and SR (success rate) [3]. However, others could arise to get more information about the performance, based on the objectives pursued by the tuning process.

In this paper we offer an alternative to perform parameter tuning of metaheuristic strategies in the design optimization of reinforced concrete (RC) frame structures, taking into account that the process of modelling, analysis and design of such structures leads to objective function evaluations that are computationally very expensive, which makes the assessment of the optimization methods very difficult. Therefore, the main goals of this paper are: (1) to propose strategies to enable parameter tuning of metaheuristics methods for (discrete) design optimization of RC frame structures using a commercial software as calculation engine; (2) to get insight of this kind of processes using a novel utility, based on the area under the average performance curve, which can allow to propose a fully automated procedure and (3) to test four metaheuristics in such (discrete) optimization problems.

The structure of this paper is as follows. In section 2 the optimization problem is formulated, the optimization methods are described, an overview of parameter tuning is developed, the novel way to measure utility is introduced, in addition to the proposed strategies to be able to afford parameter tuning in real life challenging optimization problems. Section 3 is dedicated to show and to discuss the results. Finally, conclusions are drawn and future work is proposed.

## 2. PROBLEM DEFINITION AND METHODOLOGY

Within the field of structural design optimization, the effect of the control parameters of metaheuristic algorithms has hardly been studied [4]. As mentioned, the process of modelling, analysis and design of structures is often computationally expensive, especially when a commercial software is used as calculation engine, as in our case. Thus, our methodology is based on the creation of surrogate models to be able to study optimization methods. These surrogate models are obtained by relatively simple procedures. The first one consists of the creation of databases containing the objective function values of the real model, i.e., the calculation engine is used to compute all possible solutions and by storing them and using interpolation, the surrogate model is obtained. Further analysis of this model do not need the use of the calculation engine anymore. The disadvantage of this strategy is the impossibility of evaluating highly complex structures, resulting in many variables and millions of possible solutions. The second strategy tries to overcome this shortcoming. It is less accurately, and uses benchmark functions [5] to simulate the main features of real problems, i.e. discrete optimization problems with the same number of variables (each variable with the same number of possible solutions), objective functions with the presence of many local minima and "similar" response surfaces. The advantage is that solutions are evaluated by an analytical function and not by a complex process. These two alternatives are further explained in in the following subsections.

### 2.1 Formulation of the optimization problem

Real design optimization problems are usually formulated as the minimization of objectives such as economic cost [6-10], weight [1,4] (mainly for steel structures) or environmental issues (e.g. $CO_2$ emissions or Embodied Energy) [10]. For RC structures, these optimization problems are nonlinear, i.e. objective functions depend nonlinearly on the variables, with highly nonlinear complex constraints. Additionally, objective functions values depend on many factors related to the variables, and if as in our cases, the problem is formulated to properly reflect an actual (RC) design [6-10], the response surface is very difficult to optimize, since it possesses many local minima. In this section, the main features of real variables-constraints and how they are simulated in surrogate models are explained.

#### 2.1.1 Objective functions, design variables and constraints

In this study, four cases will be used as examples. Real structures were previously optimized with respect to economic objective [8,9]. Objective functions 1 and 2 are surrogate models or databases of simple real structures shown in Fig. 2. Objective functions 3 and 4 are surrogate models of the optimization of the real structure shown in Fig. 3. Note that this structure is more complex than the first ones. The Ackley and Eggholder continuous functions are used to create the surrogate models by discretizing them. This discretization process consists of creating a solution space using some evaluations of the continuous function, generating response surfaces with the desired complexity. In Fig. 3 it can be seen how the Eggholder function generates a surrogate model that is difficult to optimize, such as the real case study. Table 1 shows the particularities of the four case studies created by using the surrogate models.

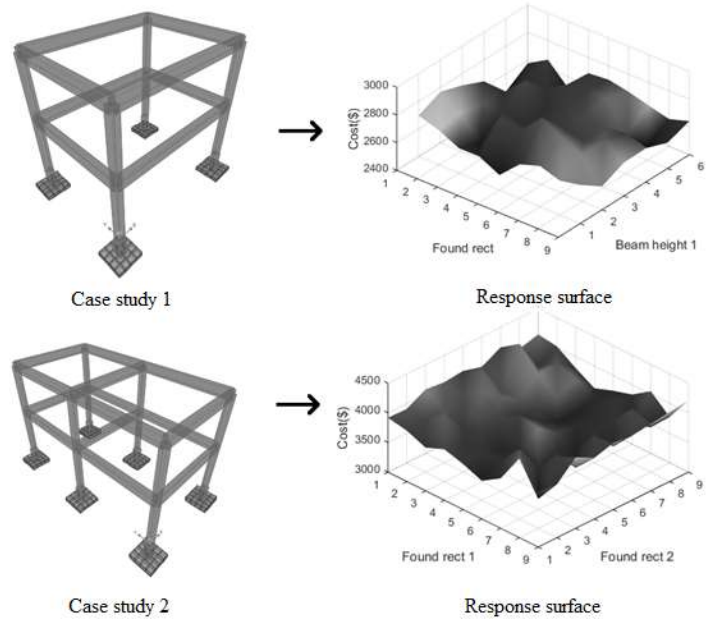

**Fig. 2** Cases studies 1 and 2 with corresponding response surfaces in function of the two most important variables. Surrogate models of these cases are obtaining by strategy 1 (storage of all possible solutions)

Real design optimization problems of RC structures usually include as variables dimensions of cross sections, solution (placement) of reinforcing steel, rectangularity of shallow foundations and properties of building materials. In our problems, these variables are discrete, in order to obtain practical solutions. This is reflected in the response surface of the objective functions, e.g., solutions that are practical from the engineering point of view, including the realistic distribution of reinforcing steel within the cross section. This introduces many local minima in the (discrete) objective function, since the configuration of steel bars varies strongly with the size of the elements. In cases 1 and 2 variables produce the same fitness value than real cases, i.e. the response surface is exactly the same as the one produced in the real problem. However, this is not true for cases 3 and 4. They are only (discrete) points obtained from the benchmark functions evaluation and fitness values belong to that function, and not to real problems. However, these points are evaluated in such a way that the response surface becomes as complicated as desired.

**Table 1** Features of the four used case studies

| Function | # variables | # possible solutions | Optimal value |
|---|---|---|---|
| Case 1 | 7 | 21870 | 2445.66 |
| Case 2 | 8 | 98460 | 3378.44 |
| Case 3 | 16 | $2.36 \times 10^{11}$ | 1.00 |
| Case 4 | 16 | $2.36 \times 10^{11}$ | -3473.20 |

In real problems, the constraints can be divided in two groups. *Design (explicit) constraints* are imposed on the design variables directly and appear for various reasons, such as functionality, manufacturing, transport or esthetic, and are of the form $X_{min} \leq X \leq X_{max}$. *Behavioral (implicit) constraints*, which are sometimes called state equations, are indirect. They deal with the fulfilment of the limit states, i.e., they define the values that the variable parameters must meet to satisfy the behavioral requirements. In our actual cases, the implicit constraints are evaluated as follows. When one of them is not satisfied, the objective function is penalized and this solution will not be selected. Consequently, these constraints influence the response surface and they are not directly taken into account in this study. However, explicit constraints are very important for surrogate models obtained by using the second strategy. They define the sub-space of the benchmark function that will be taken into account to form the surrogate model, as well as the "difficulty" of the response surface.

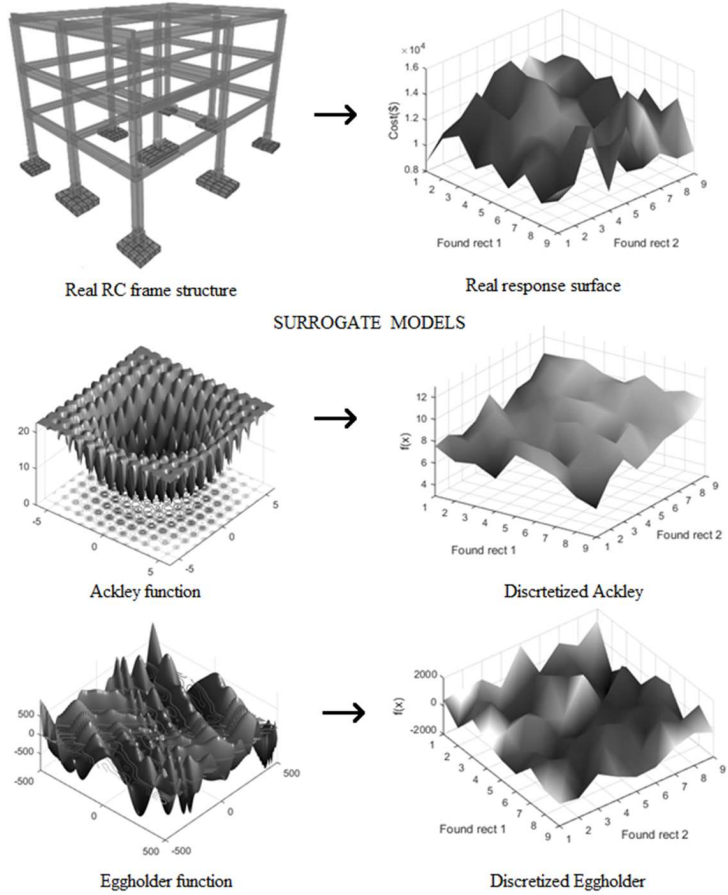

**Fig. 3** Real RC frame structure and corresponding cases studies 3 and 4, which are surrogate models based on the real structure. Response surfaces of surrogate models represent fitness values of variables simulating the rectangularity of two design groups of foundations.

## 2.2 Optimization methods

In this study, four metaheuristics are tuned and tested. Three of them are based on two of the most used ones: Genetic Algorithms (GA) and Particle Swarm Optimization (PSO). The fourth one, Biogeography-Based Optimization (BBO), is relatively new. We now briefly describe how these strategies work.

### 2.2.1  GA Matlab toolbox

The Genetic Algorithm was first proposed by John Holland [11]. GAs are a particular class of evolutionary algorithm based on the mechanics of natural selection and natural genetics. GA uses techniques inspired by evolutionary biology such as mutation, selection, and crossover.

The following outline summarizes how classical genetic algorithm works [12]. In our case, the operators work with real numbers.

1. The algorithm begins by creating a random initial population consisting of *PopSize* individuals.
2. The algorithm then creates a sequence of new populations. At each step, the algorithm uses the individuals in the current generation to create the next population. To create the new population, the algorithm performs the following steps:
    a. Scores each member of the current population by computing its fitness value.
    b. Scales the raw fitness scores to convert them into a more usable range of values.
    c. Selects members, called parents, based on their fitness. This selection process is made by using some strategies or functions (*SelFcn*).

d. Some of the individuals in the current population that have better fitness are chosen as *elite*. These elite individuals are passed to the next population. *ECountFract* specifies the percentage of the *PopSize* individuals that are guaranteed to survive to the next generation.
   e. Produces children from the parents selected by the selection function. Children are produced either by combining the vector entries of a pair of parents using a crossover function (*CrossFcn*) or by making random changes to a single parent (mutation). *CrossFract* represents the percantage of children (besides elite children) that are formed by crossover. The remaining individuals are formed by mutation.
   f. Replaces the current population with the children to form the next generation.
3. The algorithm stops when the stopping criteria is met.

### 2.2.2 GA YPEA toolbox

This variant of GA is implemented in the toolbox of evolutionary strategies YPEAv1.0 [13]. The algorithm basically works as follows:

1. It begins by creating a random initial population consisting of *PopSize* individuals, represented by a vector of real numbers.
2. The algorithm then creates a sequence of new populations following next steps:
   a. Scores each member of the current population by computing its fitness value.
   b. Selects members, called parents, based on their fitness. This selection process is made by using *roulette selection*. Here, selection pressure (*SelPress*) determines the parents' selection probabilities. Low values of *SelPress* offer a more equal selection probability among all the parents, i.e., the higher *SelPress* is, parents with best fitness will have a higher probability of being selected.
   c. This variant does not include elitism. The number of children obtained by crossover are defined by the crossover probability (*CrossProb*). Two previously selected parents *x1* and *x2* are combined to create two children *y1* and *y2* as follows:

$$y1 = (alpha * x1) + (1 - alpha) * x2 \qquad (1)$$
$$y2 = (alpha * x2) + (1 - alpha) * x1$$

   with *alpha* defined by:

$$alpha = -CrossInfl + [(1 + CrossInfl) - (-CrossInfl)] * U(0,1) \qquad (2)$$

   where *CrossInfl* is the crossover inflation or extrapolation factor and $U(0,1)$ is a random number uniformly distributed between 0 and 1.
   d. Other children are created by mutation (as long as *CrossProb* ≠ 1) as follows:

$$y = x + MutStepSize * N(0,1) \qquad (3)$$

   where *y* is the generated child, *x* is the randomly selected individual to mutate, *MutStepSize* defines the size of the change in the mutation process and $N(0,1)$ is a random number drawn from the standard normal distribution. The mutation occurs in *MutRate*MutRate**NVars* number of gens of the selected solution to mutate as in Eq. 3. *MutRate* is a fraction between 0 and 1 and *NVars* is the number of variables of the problem.

The algorithm stops when the stopping criteria is met.

### 2.2.3 PSO

Particle swarm optimization simulates the behavior of the flocking of birds, proposed by Kennedy and Eberhart [14]. The particles fly through the D-dimensional problem space by learning from the best experiences of all the particles. Therefore, the particles have a tendency to fly towards better search areas over the course of the search process.

The algorithm basically consists of the following steps:

1. The PSO algorithm begins by creating the initial particles, each with an initial position and velocity.
2. The algorithm evaluates the objective function (or fitness function) at each particle location. Then it determines the best function value and the best location.
3. It then iteratively updates the particle locations, velocities and neighbors.

a. Global information is obtained from the neighborhood. The initial neighborhood size (*N*) is obtained according to:

$$N = \max(1, floor(SwarmSize * MinFractNeigh)) \quad (4)$$

where *SwarmSize* is the number of particles and *MinFractNeigh* is the minimum adaptive neighborhood size. *N* is updated in every iteration, according to whether the best function value is improved or not.

b. The velocity of each particle is updated as follows:

$$v_{upd} = W*v + y1*u1*(p-x) + y2*u2*(g-x) \quad (5)$$

where *W* is the inertia of the particle. It is initialized with the maximum value of *InertiaRange* (or minimum if *InertiaRange* has negative values), which establishes the lower and upper bounds of the adaptive inertia; *v* is the previous velocity; *y1* (*SelfAdj*) and *y2* (*SocAdj*) are acceleration constants representing the weighting of stochastic acceleration terms that pull each particle towards the personal best and global best positions, respectively; *u1* and *u2* are uniformly distributed random numbers in [0,1]; *(p-x)* is the difference between the current position and the best position the particle has seen and *(g-x)* is the difference between the current position and the best position in the current neighborhood.

c. The position is updated according to:

$$x_{upd} = x + v_{upd} \quad (6)$$

where *x* is the previous position.

4. Iterations proceed until the algorithm reaches a stopping criterion.

### 2.2.4 Biogeography-Based Optimization

BBO is a relatively new method. It is an evolutionary algorithm (EA) and was proposed by Simon [15]. It is based on mathematical models of how species migrate from one island to another, how new species arise, and how species become extinct. Geographical areas that are well suited as residences for biological species are said to have a high habitat suitability index (HSI). The correspondence between the BBO terminology and the classical EA terminology is the following: set of habitats, habitats, species and HSI correspond to respectively population, individuals (solutions), gens and fitness value. Hence, the number of species in each habitat is equal to the number of variables in the optimization problem. The number of habitats is equal to the population size.

Since this method is rather new, it is not yet commonly used. However, its effectiveness to solve discrete problems has been shown. In [8] BBO outperformed other 10 metaheuristic techniques in the discrete design optimization of RC frame structures, including discrete GA and discrete PSO. In addition, in [9] by means of an extensive parameter tuning process, BBO demonstrated to be a very good method to solve this type of discrete optimization problems.

The algorithm consists of the following steps:

1. The procedure starts with a random initial set of habitats with a uniform HIS distribution. In every iteration, emigration and immigration coefficients, denoted by respectively μ and λ, are assigned to each habitat. Solutions or habitats with a high HSI receive high values of μ and low values of λ, and vice versa.
2. The algorithm processes the habitats in order of decreasing HSI, using the parameters μ, λ and mutation probability as follows:
   a. Within each habitat, for each species the possibility to carry out the migration process is analyzed: each species is checked based on the habitat's immigration coefficient λ. Therefore, the species in the best habitats have little chance to enter this process, while this chance increases when considering habitats with a higher λ.
   b. Once a species enters the migration process, another species from another habitat is selected using roulette wheel selection (to select the habitat) based on μ to immigrate to the habitat being worked on.
   c. Once the species are selected, immigration starts, which is not the substitution of one by the other, but a combination of both, performed as:

$$NewSpecies^i_k = Species^i_k + \alpha\,(Species^j_k - Species^i_k) \qquad (7)$$

where $Species^i_k$, i.e., the *k-th* species of habitat *i*, is the species being analyzed and $Species^j_k$, i.e., the *k-th* species of habitat *j*, is the species selected to immigrate, and $\alpha$ is the acceleration coefficient.

   d. In addition, species can mutate with a certain probability according to:

$$NewSpecies^i_k = NewSpecies^i_k + \sigma * N(0,1) \qquad (8)$$

where $\sigma$ is the mutation step size and $N(0,1)$ is a random number with mean 0 and standard deviation 1. After every iteration, $\sigma$ decreases, modified by *mutation step size damping*.

Once the entire population is analyzed, the new one is formed by selecting the best habitats of the previous population (before being transformed) and the best of the new population. *KeepRate* denotes the fraction of the previous population that survives. This is similar to elitism used in GA. This iterative process ends when a stop criterion is satisfied.

## 2.3 Parameter tuning

Parameter tuning can improve the performance of an optimization algorithm [16-18], but can be tedious and difficult to implement. Some of the most used strategies to afford this process are: Meta-Optimization, Design of Experiments (DOE), Model-Based Optimization and Machine Learning, and Model-Free Algorithm Configuration [1].

Meta-Optimization consists of using algorithmic parameters as variables in an optimization problem. Thus, the objective function is the utility metric of the algorithm performance. Depending on the problem that is generated, it can be solved by iterated local search [19] or by another metaheuristic algorithm [20-21]. The meta-optimizer can be applied either directly to the performance landscape [22] or to a surrogate model of it [23].

The most common approaches of DOE include factorial designs, coupled with analysis of variance or regression analysis [24-26]. Similar further approaches are: Design and Analysis of Computer Experiments (DACE) methodology [27], specifically created for deterministic computer algorithms [28]; Sequential Parameter Optimization (SPO) introduced by [29], which uses Kriging techniques to build a predictive model of the performance landscape; Sequential Model-based Algorithm Configuration (SMAC) introduced in [30], and [31] combines the DOE methodology with artificial neural networks as a basis for a parameter tuning framework.

Model-free Algorithm Configurations are mainly based on statistical hypothesis testing to compare different parameter settings [32-34]. The disadvantage of this method is the large number of experimental runs needed to obtain sufficient statistical accuracy. The F-Race algorithm is an efficient alternative presented in [35], where a number of predefined parameter configurations are tested on one or more benchmark problems and inferior ones are eliminated as soon as any significance arises. Balaprakash et al. [36] presented two improvement strategies for the F-Race algorithm, called sampling design and iterative refinement. Yuan and Gallagher [37] proposed a combination of F-Race and meta-optimization. López-Ibáñez [38] presented the *irace* package, offering iterated racing with a restart mechanism. Hyperband was introduced by [39] in order to adaptively allocate more resources to promising parameter configurations to enhance random search. Falkner et al. [40] combined Hyperband with Bayesian optimization in the BOHB (Bayesian optimization and Hyperband) method.

### 2.3.1 Utility metrics

To measure the performance of a metaheuristic, one must take into account the quality of the solution and the computational cost of the algorithm. Most, if not all, performance metrics used in metaheuristic optimization are based on variations and combinations of these two. Solution quality can be naturally measured by using the fitness function [3].

In [3] a summary of the most used utilities is given, stating that these measures are not always appropriate. For example, in case of large variance in the performance results of a metaheuristic, the use of the mean (and standard deviation) may not be significant and the use of median or best fitness may be preferable [41]. The performance metrics determine the optimal parameter vector. Hence, the final results of the method may vary substantially depending on the utility used. Therefore, care must be taken when defining it [42].

This strategy was initially proposed in [9]. We assume that the metaheuristic is run $N$ times with a given fixed parameter vector to reduce stochastic effects. The average performance curve, or APC (average fitness in function of the iteration number), denoted as $f_a(x)$, is obtained by averaging the performance curves of $N$ runs of the method (see Fig. 4a). This curve gives more information about performance than simple performance measures, such as MBF or AES, and allows an easy comparison between two or more procedures. Therefore, the proposed utility is based on the use of the APC.

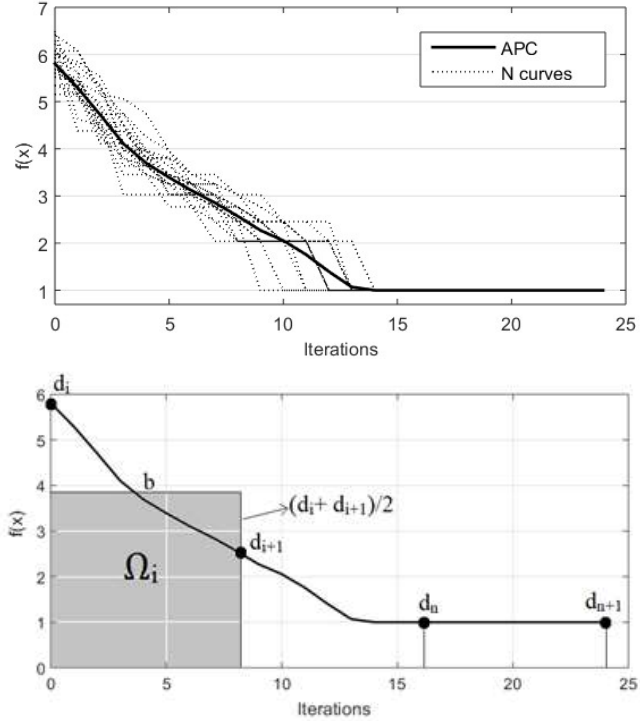

**Fig. 4** Defining the utility function. Above: the average performance curve of the $N$ curves. Below: scheme to get the utility function [9].

To assess real optimization problems, both the quality of the solution and the computational cost are important. The final point of the APC ($d_{n+1}$ in Fig. 4b) is the MBF. It only takes into account the quality of solutions. We call it $F_A$ as in Eq. 9.

$$F_A = d_{n+1} \tag{9}$$

To take into account the computational cost, the area under the average performance curve can be used. By splitting the number of iterations in $n$ equal intervals of length $b$ and by defining $d_i$, $i = 1,\ldots, n+1$ as the values of the average fitness at the end points of the intervals (Fig. 4b) we approximate the area under the average performance curve by $B$ in Eq. (10).

$$B = \sum_{i=1}^{n} \left( \frac{d_i + d_{i+1}}{2} \right) b \tag{10}$$

Note that this is equivalent to using the trapezoidal quadrature formula to approximate the integral. By means of several parameter sensitivity studies, it was decided to use $n=14$.

However, $B$ does not sufficiently take into account MBF, since it is possible that a method that converges fast to a rather high MBF has a lower $B$ than another one that converges slowly to a substantially better MBF. Hence, to ensure that MBF is properly taken into account, $F_A$ and $B$ are combined to get the proposed utility. Therefore, to make $B$ "compatible" with $F_A$, Eq. 10 is rescaled, yielding $F_B$ as in Eq. 11.

$$F_B = \frac{\sum_{i=1}^{n}\left(\frac{d_i + d_{i+1}}{2}\right)}{n} \tag{11}$$

The proposed utility $F_C$ is a weighted linear combination of utilities $F_A$ and $F_B$, with a higher weight on $F_A$, to give more significance on the solution quality than on the convergence speed, hence:

$$F_C = \frac{Z_1 * F_A + F_B}{1 + Z_1} \tag{12}$$

with $Z_1 \geq 1$, the weight that we give to utility $F_A$, i.e., MBF. We propose $Z_1 = 4$. Thus, utility $F_C$ is used to assess method's performance instead classical utilities.

### 2.4 General methodology

This methodology, as explained above, is carried out in order to gain insight into the operation of these procedures. It is based on Model-free Algorithm Configurations. In this proposal, each parameter is set with pre-established and ordered values, so that all possible combinations are tested under equal conditions. Using the $Fc$ utility in a sensitivity analysis, it was found that $N=20$ is a good value to counteract the stochastic effects of metaheuristic optimization processes, with $N$ the number of optimization procedures needed to obtain a reliable APC. Thus, if a method has four parameters and each parameter has four possible values, a total of $4^4*20$ tests will be performed to assess its performance. Table 2 introduces the parameters and parameter values that will be evaluated. As mentioned in Section 2.1 on parameter tuning, these methodologies are computationally expensive, so several strategies could be used to reduce the number of simulations needed, such as the use of Monte Carlo simulations using sampling techniques or the use of meta-models to predict response surfaces.

Two tuning processes will be proposed. The first one is very simple and consists of the selection of the configuration with the best performance in a series of similar tests (see Fig. 5a). The second one is more complicated and deals with two intermediate parameter control procedures before obtaining final results, i.e., the process starts with all possible configurations as candidates and by means of intermediate assessments, best configurations remain for further analysis (see Fig. 5b). Results will be presented by box plots where bottom whisker, box bottom, middle, top and top whisker denote minimum, 25$^{th}$ percentile, median, 75$^{th}$ percentile and maximum $Fc$ utility of each process, respectively.

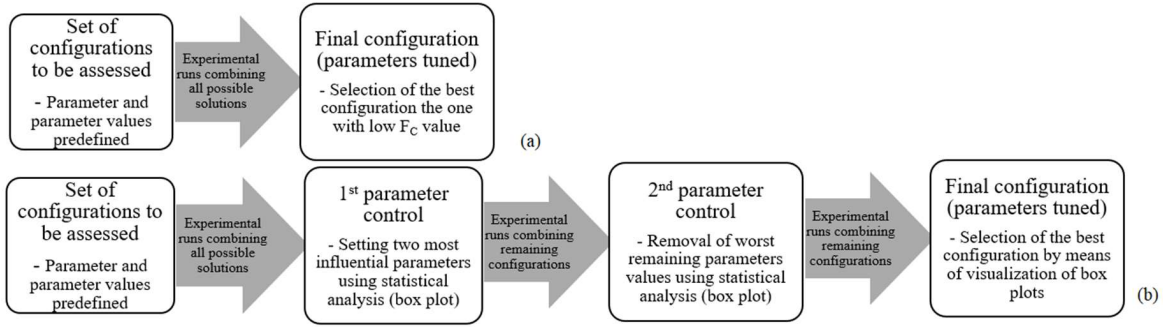

**Fig. 5** Two strategies proposed to perform parameter tuning

## 3. RESULTS AND DISCUSSIONS

The results are divided in two main points. The first one is related to comparisons between methods' performances for the given optimization problems. The second one is about finding the way of getting an efficient strategy to perform parameter tuning for linked problems. This would be the first step to create a fully automated tuning procedure based on concepts developed in this work.

### 3.1 Comparison between the four methods

In this section, the four methods are assessed and compared. This comparison is made by performing and storing $F_C$ values of all the possible parameter and parameter value combinations of each method. Tables 3, 4, 5 and 6 show the configurations of the three best $F_C$ values obtained by each method tested in each case study. Some conclusions can be drawn about best configurations of each method to solve discrete optimization problems, e.g. *tournament* selection function seems to be the best option for strategies based on GA, or high ratios of *social adjustment* for swarm intelligence algorithms. How these parameters influence the methods' behavior can be found in section 2.2

**Table 2** Parameter and parameter values to be tuned

| Method | Parameter | Parameter values |
|---|---|---|
| GA toolbox | PopSize | 50; 100; 150; 200 |
| | ECountFract | 0.05; 0.10; 0.15; 0.20 |
| | CrossFract | 0.70; 0.80; 0.90; 1.00 |
| | SelFcn | "stochunif"; "remainder"; "uniform"; "roulette"; "tournament" |
| | CrossFcn | "scattered"; "intermediate"; "heuristic"; "sinpoint"; "twopoints"; "arithmetic" |
| GA YPEA | PopSize | 50; 100; 150; 200 |
| | CrossProb | 0.60; 0.70; 0.80; 0.90 |
| | CrossInfl | 0.10; 0.20; 0.30; 0.40 |
| | MutRate | 0.10; 0.20; 0.30; 0.40 |
| | MutStepSize | 0.05; 0.10; 0.15; 0.20 |
| | SelPressure | 1; 3; 5 |
| PSO | SwarmSize | 50; 100; 150; 200 |
| | MinFractNeigh | 0.10; 0.20; 0.30; 0.40 |
| | SelfAdj | 0.50; 1.00; 1.49; 1.99 |
| | SocialAdj | 0.50; 1.00; 1.49; 1.99 |
| BBO | PopSize | 50; 80; 100; 120; 140 |
| | Alpha | 0.90; 0.95; 0.99 |
| | MutProb | 0.30; 0.40; 0.50 |
| | MutStepSize | 0.025; 0.050; 0.075; 0.100 |
| | MutStepSizeDamp | 0.99; 1.00; 1.01; 1.02 |

**Table 3** Three best $F_C$ values obtained using GA-toolbox method

| Function | Best values | PopSize | ECountFract | CrossFract | SelFcn | CrossFcn | $F_C$ value |
|---|---|---|---|---|---|---|---|
| Case 4A | 1 | 200 | 0.10 | 0.90 | Remainder | Scattered | 2449.989 |
| | 2 | 200 | 0.15 | 0.90 | Tournament | Scattered | 2450.066 |
| | 3 | 200 | 0.10 | 0.90 | Tournament | Scattered | 2450.245 |
| Case 4B | 1 | 200 | 0.10 | 0.90 | Tournament | Scattered | 3383.426 |
| | 2 | 200 | 0.10 | 1.00 | Tournament | Scattered | 3384.222 |
| | 3 | 200 | 0.05 | 0.80 | Tournament | Scattered | 3384.249 |
| Ackley | 1 | 150 | 0.00 | 0.90 | Tournament | Scattered | 1.319 |
| | 2 | 200 | 0.00 | 1.00 | Tournament | Scattered | 1.355 |
| | 3 | 200 | 0.00 | 0.90 | Tournament | Scattered | 1.379 |
| Eggholder | 1 | 200 | 0.00 | 1.00 | Tournament | Single Point | -3125.725 |
| | 2 | 200 | 0.15 | 1.00 | Tournament | Scattered | -3110.837 |
| | 3 | 200 | 0.00 | 0.80 | Tournament | Two Point | -3098.315 |

Tables 3-6 and Fig. 6 show that cases 1 and 2 are relatively the solved by the four methods. However, cases 3 and 4 are much more difficult to optimize, and GA-toolbox and PSO perform worse compared to GA in YPEA Toolbox and BBO. The last one has excellent results while optimizing all case studies, also showing that is a good "generalist", i.e., it offers good results regardless of the used configuration.

The first and simplest strategy proposed to perform parameter tuning (Fig. 5a) consists of selecting as the best configuration the one with the best $F_C$ (bottom whisker in box plots). The main disadvantage is that a deep statistical analysis is not considered, and configurations with faster convergence to any optimum (local or global) may have good $F_C$ values, and yet, the global one cannot frequently be found. Other configurations with lower FC values may take longer to converge, but the global optimum is found more frequently. This strategy will be called "parameter tuning 1". Other alternatives can include the use of the median or the mean. However, in simple procedures such as the proposed one, the complex interaction between parameters and parameter values is not taken into account, i.e., one parameter value can lead to excellent results with some specific values of other parameters and to very poor results with other ones, affecting the mean or the median, thus this parameter value could be neglected even when, in other circumstances, it could be the best, or a good one at least.

**Table 4** Three best $F_C$ values obtained using GA-YPEA method

| Function | Best values | PopSize | CrossProb | CrossInfl | MutRate | MutStepSize | SelPress | $F_C$ value |
|---|---|---|---|---|---|---|---|---|
| Case 4A | 1 | 50 | 0.60 | 0.30 | 0.30 | 0.20 | 3 | 2447.464 |
| | 2 | 50 | 0.60 | 0.30 | 0.40 | 0.20 | 3 | 2447.618 |
| | 3 | 50 | 0.60 | 0.20 | 0.30 | 0.20 | 5 | 2447.677 |
| Case 4B | 1 | 50 | 0.60 | 0.30 | 0.40 | 0.20 | 1 | 3381.301 |
| | 2 | 50 | 0.60 | 0.30 | 0.40 | 0.20 | 5 | 3381.316 |
| | 3 | 50 | 0.80 | 0.30 | 0.30 | 0.20 | 1 | 3381.368 |
| Ackley | 1 | 50 | 0.60 | 0.30 | 0.20 | 0.20 | 1 | 1.196 |
| | 2 | 50 | 0.70 | 0.30 | 0.10 | 0.20 | 1 | 1.199 |
| | 3 | 50 | 0.60 | 0.30 | 0.40 | 0.15 | 1 | 1.207 |
| Eggholder | 1 | 50 | 0.60 | 0.30 | 0.40 | 0.20 | 5 | -3022.268 |
| | 2 | 50 | 0.60 | 0.30 | 0.40 | 0.20 | 3 | -3022.228 |
| | 3 | 50 | 0.60 | 0.30 | 0.30 | 0.20 | 3 | -3016.860 |

**Table 5** Three best $F_C$ values obtained using PSO method

| Function | Best values | SwarmSize | MinFractNeigh | SelfAdj | SocialAdj | $F_C$ value |
|---|---|---|---|---|---|---|
| Case 4A | 1 | 200 | 0.30 | 1.49 | 1.99 | 2446.003 |
| | 2 | 150 | 0.40 | 1.99 | 1.99 | 2446.053 |
| | 3 | 200 | 0.20 | 1.99 | 1.99 | 2446.104 |
| Case 4B | 1 | 100 | 0.30 | 1.99 | 1.49 | 3380.028 |
| | 2 | 150 | 0.30 | 1.49 | 1.99 | 3380.111 |
| | 3 | 100 | 0.10 | 1.99 | 1.99 | 3380.336 |
| Ackley | 1 | 50 | 0.30 | 1.49 | 1.99 | 2.645 |
| | 2 | 50 | 0.20 | 1.99 | 1.99 | 2.697 |
| | 3 | 50 | 0.20 | 1.49 | 1.99 | 2.737 |
| Eggholder | 1 | 50 | 0.20 | 1.99 | 1.99 | -3025.778 |
| | 2 | 50 | 0.10 | 1.49 | 1.99 | -3019.160 |
| | 3 | 50 | 0.10 | 1.99 | 1.99 | -3013.728 |

**Table 6** Three best $F_C$ values obtained using BBO method

| Function | Best values | PopSize | Alpha | MutProb | MutStepSize | MutStepSizeDamp | $F_C$ value |
|---|---|---|---|---|---|---|---|
| Case 4A | 1 | 60 | 0.99 | 0.50 | 0.10 | 1.02 | 2446.694 |
| | 2 | 60 | 0.99 | 0.40 | 0.10 | 0.99 | 2446.716 |
| | 3 | 60 | 0.95 | 0.40 | 0.10 | 1.00 | 2446.720 |
| Case 4B | 1 | 60 | 0.99 | 0.50 | 0.075 | 1.01 | 3379.987 |
| | 2 | 60 | 0.99 | 0.30 | 0.10 | 1.01 | 3380.044 |
| | 3 | 60 | 0.99 | 0.30 | 0.075 | 1.01 | 3380.046 |
| Ackley | 1 | 60 | 0.99 | 0.30 | 0.05 | 0.99 | 1.137 |
| | 2 | 60 | 0.99 | 0.30 | 0.05 | 1.02 | 1.138 |
| | 3 | 60 | 0.95 | 0.50 | 0.05 | 1.00 | 1.139 |
| Eggholder | 1 | 80 | 0.99 | 0.50 | 0.075 | 1.01 | -3218.605 |
| | 2 | 80 | 0.99 | 0.30 | 0.075 | 1.01 | -3215.875 |
| | 3 | 80 | 0.99 | 0.50 | 0.075 | 0.99 | -3211.563 |

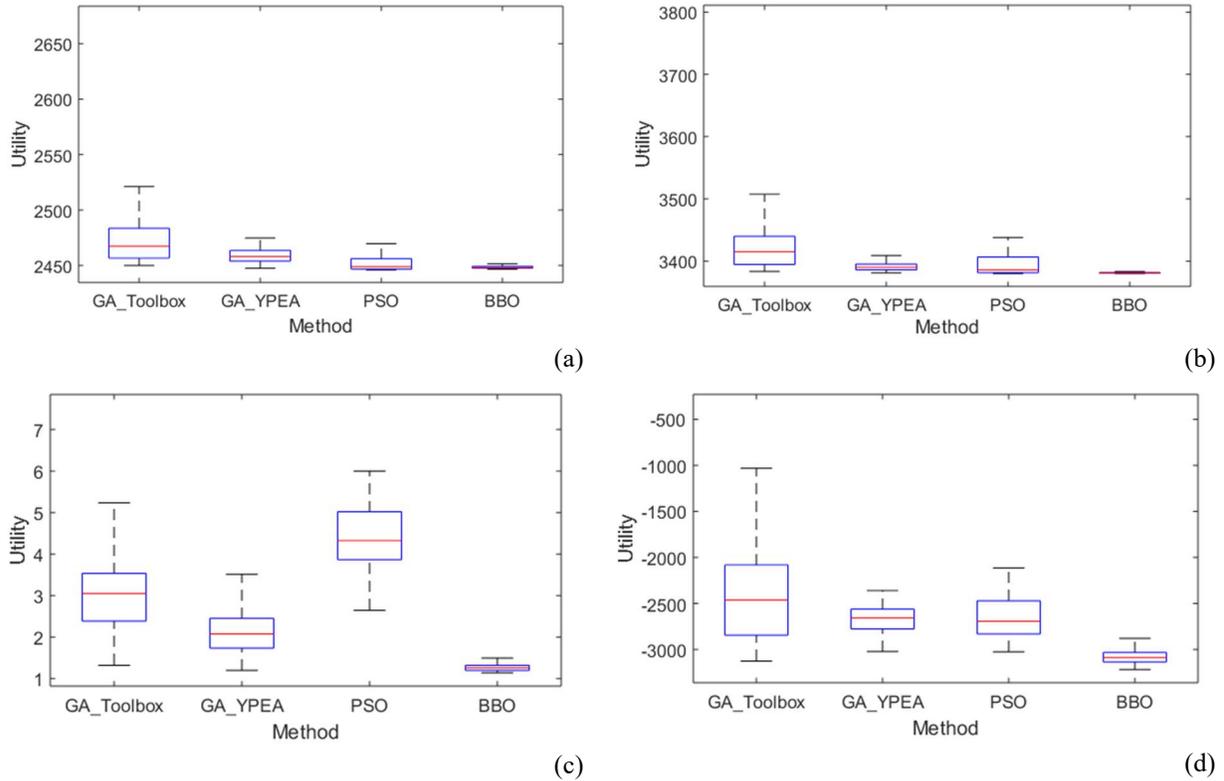

**Fig. 6** Summary of results of the four methods applied to the four case studies using box plots with $F_C$ values; (a), (b), (c) and (d) represent case studies 1, 2, 3, and 4 respectively.

### 3.2 Use of parameter control to tune the parameters

In order to avoid the last shortcoming, another strategy will be developed based on the concept of parameter control (see Fig. 5b). As explained, the process has two phases. In the first one, instead of finalizing the process, a new set of parameter is fixed as follows: the two most evident influential parameters (we consider a parameter to be influential if varying its value leads to substantial differences in performance) are fixed with the best values. A number of new simulations are made based on the remaining possible configurations. Using this new result (some remaining parameters may change the "optimal value" obtained in the first series of test), in the second phase parameter values with an "evident" poor performance, are omitted, and with this new configuration, the final number of simulations are run. This new result give the final solution. This will be called "parameter tuning 2".

Results after applying the first phase are shown in Fig. 7. It was decided to keep PopSize as a variable parameter even when it is very significant, due to its strong dependence on the values of other parameters and its importance as a regulator of solution quality and computational cost. It can be seen how, in both cases, results substantially improve (except GA YPEA applied to the case 4).

In this first phase, GA YPEA gives significant performance improvements in cases 1, 2 and 3. However, the enhancement in case 4 is not substantial. This means that, as this is the most difficult function to optimize, there are still parameter values with poor performance.

However, the improvements in BBO performance is quite significant in all cases. It is because the set of parameters is very influential on the method's behavior. E.g., it seems that Alpha, which determines the magnitude of the immigration process (or combination between the two analyzed variables, see Eq. 7), is quite significant for the recombination operator of BBO.

The importance of this first phase is not only checked by the mentioned improvements, but also by changes of ostensible optimal parameter values result of the first series of tests, e.g. *MutStepSize* (case 2) and *CrossProb* (case 4) from GA YPEA or *MutProb* (case 2) and *PopSize* (case 4) from BBO.

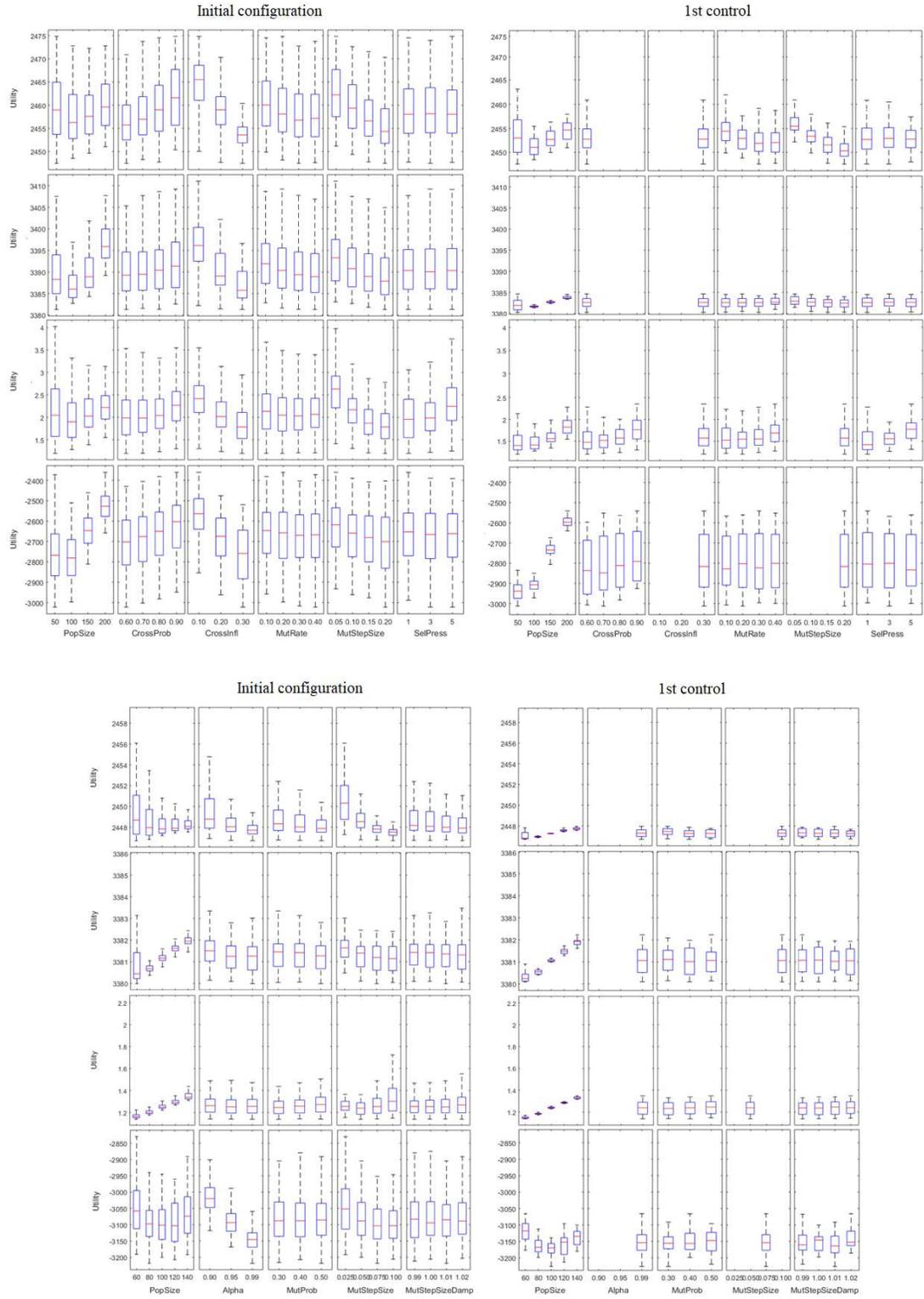

Fig. 7 Result of applying the first control. Above: GA YPEA applied to the four cases studies. Below: BBO also applied to the four cases. Note that y-axis ranges are not the same for both methods, so comparisons between methods are not allowable.

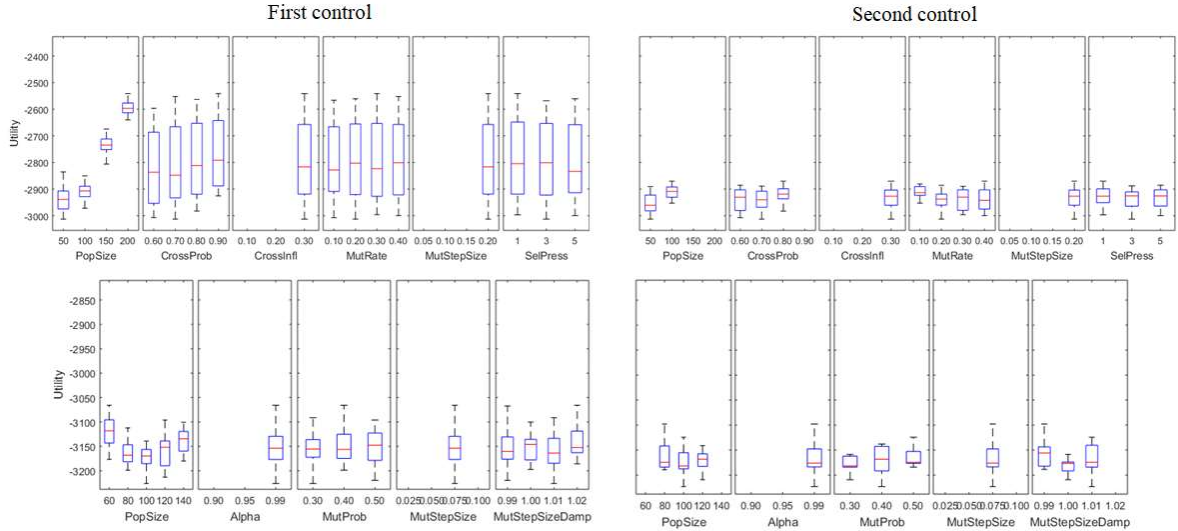

**Fig. 8** Result of applying the second control. Above: GA YPEA applied to case study 4. Below, BBO also applied to case study 4. Note that y-axis ranges are not the same for both methods, so comparisons between methods are not allowable.

The second phase was only applied to the case study 4. In Fig. 8, it can be seen that the improvements achieved with GA YPEA are outstanding, obtained by the elimination of some parameter values such as population sizes of 150 and 200. BBO does not achieve significant enhancements, due to the high accuracy of the first phase. The most important changes are the improvement of values such as population size of 100, 40% mutation probability and no damping of the mutation step size. A summary of the results of each configuration is shown in Fig. 9.

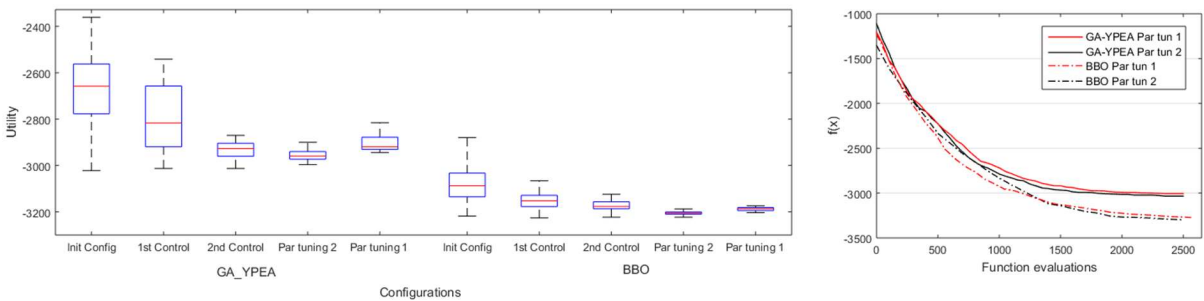

**Fig. 9** Summary of results. Left: comparison between all configurations, parameter tuning 1 is the first strategy proposed and parameter tuning 2 is the second one including initial configuration, first and second control results to assess methods' performance. Right: APCs of two tuning strategies for GA YPEA and BBO.

This figure shows the progressive improvement in the performance of the methods and the superiority of the second tuning strategy. In order to assess the performance of the two tuning strategies, 50 independent runs were performed using each configuration resulting from both tuning processes. For GA YPEA, the second strategy clearly outperformed the first one, which is not the case of BBO. It is important to take into consideration the property of good "generalist" of this method, i.e., it is not difficult to find a "good configuration" to deal with similar problems.

On the other hand, the APCs shown in this figure demonstrate the efficiency of the proposed utility. The performances of the curves are similar, especially the ones belonging to the same method. However, results shown using box plots (which use $F_C$ values) clarify the results. The case of BBO is interesting. The curve belonging to strategy 1 has a faster convergence to good solutions. The area under this curve is smaller than the one belonging to the strategy 2. However, the last one finally converges to better results, resulting in a better MBF. This is why box plots of tuning strategies using BBO are quite similar. If $F_C$ would not assign a higher weight to MBF than to the area under the APC, the implied box plots would probably indicate the opposite result. Otherwise, if the MBF is used instead of $F_C$, box plots would indicate that the difference between the two tuning strategies is larger, which is not the case. When analyzing the results achieved with GA YPEA, the two APCs are almost the same. Nevertheless, box plots suggest that the second strategy is clearly better than the first one. Therefore, the proposed utility $F_C$ takes into account more information than classical utility metrics.

Finally, it would be interesting to analyze the good performance of BBO when dealing with this type of discrete optimization problems, evidently superior to the other methods. The great difference between BBO and classical related metaheuristics, such as GA, is the recombination and mutation operator used. Classical EAs combine entire solutions, i.e., two previous solutions are selected as parents (in case of GA), and they are "combined" to obtain the new solution, or child. This means that the analysis is done from solution to solution. In the case of BBO, this process of combining solutions is done from variable to variable. This means that solutions are formed by a number of variables. BBO performs this analysis from variable to variable (or species in the BBO terminology), and when GA combine two solutions to create a new one, BBO can obtain solutions from more than two previous candidates. In addition, the combination (Eq. 8) and the mutation (Eq. 9) operators, in the same process of getting new solutions, can affect the variable involved. This could be the explanation of why BBO seems to be a better strategy to solve discrete optimization problems.

## 4. CONCLUDING REMARKS AND FUTURE WORK

In this paper, alternatives to tune the stochastic operators and their parameters of metaheuristic methods are presented. We consider genetic algorithms, particle swarm optimization and biogeography-based optimization applied to the structural design optimization of reinforced concrete frame structures. An important issue is the computational cost of these metaheuristic methods, especially when they use a commercial software as calculation engine to determine the fitness or objective function. Since tuning processes themselves are also expensive, tuning the parameters for such optimization problems is very challenging. Therefore we propose the use of surrogate models instead of the real models to evaluate the fitness or objective function. Another novel feature introduced in this paper is the use of a utility metric based on the average performance curve (APC), instead of classical ones such as mean best fitness, average number of evaluations to solution or success rate. This new utility metric offers more information about the performance of the metaheuristics.

Further, two alternatives to perform parameter tuning are presented. Ideally, all possible combinations of parameters and parameter values should be evaluated. The first one is very simple and consists of running experiments with many (or all) parameter and parameter values and select the best one as the tuned parameter configuration. The second one is more complex and requires more computations. It takes into account the concept of parameter control, and consists of two phases, in which initial parameters and parameter values change according to statistical analysis done with the intermediate results. The first strategy offers good results only for methods that are good generalists, i.e., methods that offer good performance regardless of the used parameter configuration. The second strategy leads to much better results.

It is important to highlight that biogeography-based optimization (BBO) is able to deal very well with discrete optimization problems, due to its particular strategy to obtain the new solutions. Classical EAs create, independently, new solutions by means of crossover of two (or more) individuals, mutation or elitism. BBO is able to create a new solution by recombination and mutation of variables in several individuals.

Nevertheless, the proposed methodology still has two disadvantages. Even when the computational cost is substantially reduced by the use of surrogate models, such models are only suitable for simple structures (first type of surrogate models) or are not sufficiently analogous to the real ones (second type). The use of meta-models to be able to create more accurate models could be a possible solution for this shortcoming. The other disadvantage is related to the tuning process, which needs many runs to be able to tune the parameters. Further work will use strategies such as Monte Carlo simulations to combine parameters and parameter values in a more efficient way. Additionally, the use of meta-models will be used to predict the utility landscape, also reducing the number of experimental runs needed.